# Self-Supervised Relevance Modelling in Autonomous Driving via Counterfactual Analysis

Luca Lusvarghi, Javier Gozalvez, Pablo Urbano Hidalgo
Networked Systems Lab, Universidad Miguel Hernandez de Elche, Elche, Spain.
Email: {llusvarghi, j.gozalvez, purbano}@umh.es

***Abstract*—Autonomous driving relies on computationally intensive perception pipelines to continuously detect and track objects in the surrounding environment. While some objects are key to plan safe and effective maneuvers, others may not be relevant and have no impact on the autonomous vehicle's driving decisions. Focusing on relevant objects allows a more efficient usage of available computational resources, reduces processing latencies, and limits the downstream propagation of perception noise. In this work, we propose a novel self-supervised approach based on counterfactual analysis to develop a relevance model – an AI-based tool that quantifies the relevance of objects for an autonomous vehicle. To demonstrate the potential of the proposed approach, we train a relevance model on a synthetic causal dataset generated in a selected urban scenario. Results show that the relevance model is able to accurately estimate the objects' relevance with millisecond-level latency, enabling real-time relevance estimation also in high-density scenarios. We also show that the relevance model can be used to build relevance heatmaps that offer valuable insights into the autonomous vehicle's driving policy and can be used to proactively inform perception and planning tasks. We openly release both the relevance model and the causal dataset.**

***Keywords—autonomous driving, relevance, counterfactual analysis, causal learning, self-supervised, perception***

## I. Introduction

Autonomous driving systems rely on the continuous perception of the surrounding environment to make safe and effective driving decisions. Computationally intensive onboard perception pipelines integrate data from heterogenous sensors – such as cameras, LiDARs, and radars – to detect and track objects (e.g., vehicles, pedestrians, traffic cones) within the ego-vehicle's surroundings. While some objects are critical for driving safety and compliance with the traffic rules, others may not have any impact on the ego-vehicle's driving decisions. Understanding the objects' relevance is therefore key to optimize the usage of computational resources, particularly in high-density scenarios where the number of detected objects can be large. Focusing on relevant objects allows an autonomous vehicle to relax the computational strain on its hardware (CPUs, GPUs), ultimately lowering onboard processing latencies and energy consumption. Moreover, it reduces the noise injected into downstream planning tasks, allowing the autonomous vehicle to make more effective driving decisions, as highlighted in [1].

A number of research works have recently attempted to model the relevance of surrounding information by focusing on the driver's attention. In [2], the authors presented an AI-based Focus of Attention (FoA) prediction model trained on real-world datasets. The FoA corresponds to the field-of-view region that a human driver pays attention to while driving and, therefore, is supposed to contain relevant information influencing the driver's decisions. More recent contributions extended the approach proposed in [2] by analyzing the impact of driver's intentions (turn left or right, go straight) [3], driving experience and route familiarity [4] on the FoA prediction accuracy. The attention of an autonomous agent, instead of a human driver, was analyzed in [5] to identify the most relevant object influencing the learned driving policy. Another line of research works investigating the relevance of detected objects is framed within the Object Importance Estimation (OIE) context [6]-[9]. In [6] and [7], OIE is addressed through AI-based binary classification models trained on human-annotated datasets of real-world driving videos. In [8] and [9], object importance is binarily classified leveraging counterfactual analysis. Counterfactual analysis is a powerful causal learning tool that, unlike traditional AI, enables the extraction of causal cues representing explicit causal relationships rather than mere statistical correlations. It involves analyzing a system's behavior under alternative counterfactual scenarios obtained by modifying a system variable while keeping all others constant. Counterfactual scenarios generated by iteratively removing detected objects from the scene – or by changing their speed – are leveraged to evaluate the impact of each object on the ego-vehicle's behavior in both [8] and [9]. The ego-vehicle's behavior is approximated using a trajectory prediction model, and the OIE accuracy is evaluated on human-annotated datasets of synthetic CARLA [10] scenarios and real-world driving videos in [8] and [9], respectively. In both [6]-[7] and [8]-[9], dataset annotations are performed offline and an object is labelled as important if it could potentially influence the ego-vehicle's driving according to the annotator. Human-annotated datasets are poorly scalable due to their high cost (in terms of time and economic resources) and are generated relying on the annotator's common sense, which may not accurately capture an autonomous vehicle's driving needs.

In this work, we propose a novel self-supervised approach based on counterfactual analysis to model the relevance of objects for an autonomous vehicle. In the proposed approach, counterfactual analysis is not employed to binarily classify objects [8]. Instead, it is used to learn. Causal cues linking the relevance of an object with the ego-vehicle's state (position, heading, speed, future intentions) are extracted through counterfactual analysis and encoded into a causal dataset. The causal dataset is then leveraged to train an AI-based relevance model – a tool that captures (i.e., learns) the direct causal relationships that determine each object's relevance, rather than statistical correlations. Unlike existing works, we don't define relevance as a binary concept (relevant/irrelevant), nor do we rely on human annotations. In the proposed approach, the relevance of an object quantifies its impact on the driving decisions of an autonomous vehicle, and is determined by directly evaluating the object's impact on the autonomous vehicle's planned trajectory through counterfactual analysis. Counterfactual scenarios are generated by selectively adding objects rather than removing them. We show that this design choice allows to eliminate confounding effects that may emerge from the presence of other objects in the scenario and compromise the relevance estimation accuracy. According to relevance's non-binary definition, the relevance model is

This work was supported by the European Union under the 2023 MSCA Postdoctoral Fellowship program (project no. 101153845) and by MCIN/AEI/10.13039/501100011033 (PID2023-150308OB-I00).

designed as a regression task and it is trained leveraging the objects' relevance as the only supervision signal. The self-supervised nature of the proposed approach makes it significantly more scalable with respect to state-of-the-art proposals that rely on human supervision to assess object importance (or relevance).

We demonstrate the potential of the proposed approach in a selected urban scenario generated using SUMO [11], considering an ego-vehicle running Autoware's open-source autonomous driving stack [12]. The obtained results show that (i) the proposed approach allows to more accurately determine the objects' relevance compared to existing approaches that rely on human annotations and (ii) the learned relevance model can accurately estimate the relevance of an object with millisecond-level inference latency on a single CPU. The proposed relevance model outperforms existing online counterfactual analysis approaches [8][9] in terms of inference latency and enables scalable real-time relevance estimation also in high-density scenarios with several objects. Furthermore, we show that the relevance model can be used to build relevance heatmaps. Relevance heatmaps offer a spatial understanding of the objects' relevance, provide valuable insights into the autonomous vehicle's driving policy, and can allow autonomous vehicles to proactively focus perception and planning efforts on specific areas of the surrounding environment. We openly release both the relevance model and the causal dataset presented in this work at [13].

## II. Causal Dataset

A causal dataset is essential for developing an accurate AI-based relevance model that efficiently learns true causal relationships rather than statistical correlations. We generate the causal dataset in a self-supervised manner, directly measuring the objects' relevance for an Autoware-based ego-vehicle through counterfactual analysis on synthetic driving scenarios generated using SUMO.

### A. Driving Scenario Generation

We define a driving scenario as a snapshot taken at time $t$ from a road network (with lanes, driving rules, road priorities, etc.) populated by the ego-vehicle and a set of surrounding objects. The ego-vehicle's state is defined by its position $(x_e, y_e)$, heading $(\theta_e)$, speed $(v_e)$, and goal destination $(x_{g,e}, y_{g,e})$. The ego-vehicle's position on the road network is randomly sampled and represents the center of its coordinates system. Its heading is defined in accordance with the driving rules (left- or right-hand traffic) that characterize the ego-vehicle's current lane. Its speed is randomly sampled in the $[0, v_{max}]$ m/s range, where $v_{max}$ denotes the speed limit on the current lane. The goal destination is randomly sampled so that its driving distance with respect to the ego-vehicle's position is larger than $D_{goal}$ m. The goal destination represents the final position that the ego-vehicle aims to reach and, therefore, influences its future driving intentions (turn left or right, go straight).

At time $t$, we extract ground-truth information from all objects within a driving distance range $D_{max}$ from the ego-vehicle. $D_{max}$ is defined to exclude from the causal dataset all objects that are too far from the ego-vehicle and cannot have any impact on its driving decisions. The objects mobility is simulated with SUMO. Let's indicate with $N$ the total number of objects in the driving scenario that lie within the $D_{max}$ range, and with $o_n$ the $n$-th object $(n = 1, .., N)$. The ground-truth information extracted from $o_n$ at time $t$ is the following: object's width $(W_n)$ and length $(L_n)$, class $(c_n)$, position $(x_n, y_n)$, heading $(\theta_n)$, speed $(v_n)$, and predicted trajectory $(\varphi_n^T)$. The class attribute represents the object type (vehicle, pedestrian, truck, etc.). Position and heading refer to the center of its bounding box in SUMO. Each object's ground-truth predicted trajectory represents the object's future position over the next $T_n$ seconds and is obtained by advancing the SUMO simulation beyond time $t$. According to Autoware's requirements, the length of each object's predicted trajectory is $T_n$ = 10 seconds and its sampling period is $\Delta t_n$ = 500 milliseconds.

Intentionally, the set of extracted objects is not restricted to those detected by the ego-vehicle through its onboard sensors to ensure that the generated causal dataset is independent of the ego-vehicle's sensing capabilities.

### B. Counterfactual Analysis

In the proposed approach, the relevance of an object $o_n$ quantifies its impact on the ego-vehicle's driving decisions, in particular, on its planned trajectory. In autonomous driving, the planned trajectory consists of a time-parameterized path that defines the short-term evolution of the ego-vehicle's state over time:

$$\psi_e^T = \{(x_e, y_e, \theta_e, v_e)_\tau\}_{\tau=t+\Delta t}^{T}, \quad (1)$$

where $T$ and $\Delta t$ represent the planned trajectory's duration and sampling period, respectively. In Autoware [12], $T$ = 5 seconds and $\Delta t$ = 100 milliseconds.

We extract the relevance of an object $o_n$ through counterfactual analysis. Counterfactual analysis is performed by evaluating the difference between $\psi_{e,ref}^T$ and $\psi_{e,n}^T$. $\psi_{e,ref}^T$ represents the reference trajectory planned by the ego-vehicle in an empty reference scenario, assuming that there are no objects in the surrounding environment. It represents the optimal trajectory that the ego-vehicle can plan to drive towards its goal destination. $\psi_{e,n}^T$ represents the trajectory planned by the ego-vehicle in a counterfactual scenario populated by only $o_n$. We obtain $\psi_{e,n}^T$ by directly injecting $o_n$'s ground-truth object-level information extracted from SUMO into the ego-vehicle's downstream planning modules. The relevance of an object $o_n$ corresponds to the difference between $\psi_{e,ref}^T$ and $\psi_{e,n}^T$ and is measured through the Average Displacement Error (ADE). For each object $o_n$, $\text{ADE}_n$ measures the average Euclidean distance between the ego-vehicle's positions in $\psi_{e,ref}^T$ and $\psi_{e,n}^T$, quantifying the object's impact on the ego-vehicle's planned trajectory:

$$\text{ADE}_n = \frac{1}{\text{T}} \sum_{\tau=t+\Delta t}^{T} \left\| \left(x_{e,ref}, y_{e,ref}\right)_\tau - \left(x_{e,n}, y_{e,n}\right)_\tau \right\|_2. \quad (2)$$

Unlike existing approaches [8][9], we don't perform counterfactual analysis by removing objects from the driving scenario. Instead, we selectively add each object to an empty reference scenario. The proposed approach allows to more accurately capture the causal cues that determine each object's relevance, removing all potential confounders. In causal learning, a confounder is a system variable that creates a spurious association between the observed quantities, compromising a correct estimation of their causal relationship.

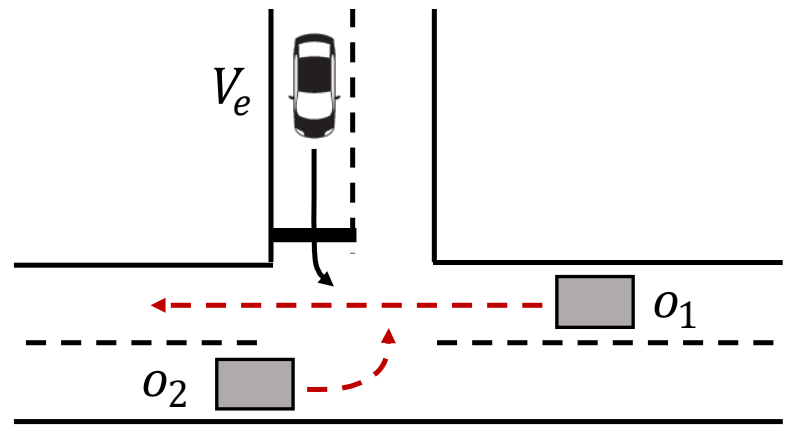

Fig. 1. Driving scenario example. Objects correspond to vehicles and are represented by their bounding box.

We illustrate the effectiveness of the proposed approach considering the T-shaped intersection depicted in Fig. 1, where the ego-vehicle $V_e$ plans to turn left but stops and yields to two approaching objects, $o_1$ and $o_2$. Let's assume that both $o_1$ and $o_2$ are vehicles. Existing approaches measure the relevance of an object $o_n$ by analyzing the impact of its removal on the ego-vehicle's behavior, and would estimate that neither $o_1$ nor $o_2$ are relevant for $V_e$ in Fig. 1. When removing $o_1$, $V_e$ would still not be able to cross the intersection and turn left due to the presence of $o_2$. Therefore, its behaviour would not change and $o_1$ would be considered irrelevant. Vice versa, $V_e$'s behaviour would not change when removing $o_2$, due to presence of $o_1$. In this example, the two objects are confounding each other's relevance. In contrast, the approach proposed in this work evaluates the relevance of each object by comparing its impact against an empty reference scenario with no confounders. In the empty reference scenario, $V_e$ can cross the intersection and turn left, as there are no surrounding objects potentially affecting its driving decisions. By selectively adding each object to the empty scenario, the counterfactual analysis would reveal that both $o_1$ and $o_2$ affect the ego-vehicle's behavior (i.e., its planned trajectory), forcing the ego-vehicle to stop. The proposed approach would then correctly label both object $o_1$ and $o_2$ as relevant.

We iteratively perform the counterfactual analysis on all $N$ objects collected in each driving scenario, and we repeat it across a large number of driving scenarios to obtain a sufficiently rich causal dataset. For each object $o_n$, we encode the following information in the causal dataset:

- Ego-vehicle's state: position $(x_e, y_e)$, heading $(\theta_e)$, speed $(v_e)$, and goal destination $(x_{g,e}, y_{g,e})$.
- Object's dimensions $(W_n, L_n)$, class $(c_n)$, position $(x_n, y_n)$, heading $(\theta_n)$, speed $(v_n)$, predicted trajectory $(\varphi_n^T)$, and ADE$_n$ (i.e., its relevance).

We integrate the causal dataset with map-level information about the road network. Note that the approach proposed in this work to generate the causal dataset is not specific to Autoware and can be adapted to other autonomous driving stacks.

## III. Modelling Relevance

This section presents the first relevance model for autonomous driving derived with the proposed approach. The relevance model has been derived considering a T-shaped urban intersection scenario and an ego-vehicle running Autoware's open-source autonomous driving stack [12]. The selected intersection is extracted from CARLA's Town01 map and illustrated in Fig. 2. It does not have traffic lights, it has one lane per driving direction, and is populated only by vehicles. The maximum speed on all lanes is $v_{max}$ = 50 km/h. In this setting, we configure the $D_{goal}$ and $D_{max}$ ranges to 200 m and 300 m, respectively. The $D_{goal}$ setting guarantees that the ego-vehicle can always plan a full $T$ = 5 seconds-long planned trajectory, even when travelling at the maximum allowed speed $v_{max}$. The $D_{max}$ setting excludes from the causal dataset all objects whose predicted trajectory cannot intersect the ego-vehicle's planned trajectory even when their relative speed is maximum (i.e., equal to $2 \cdot v_{max}$).

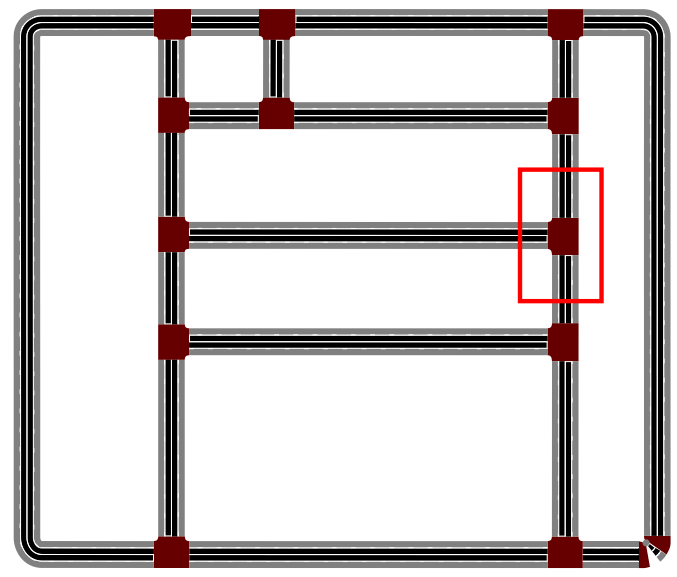

Fig. 2. CARLA's Town01 map road network. The T-shaped intersection considered in this work is highlighted by the red box.

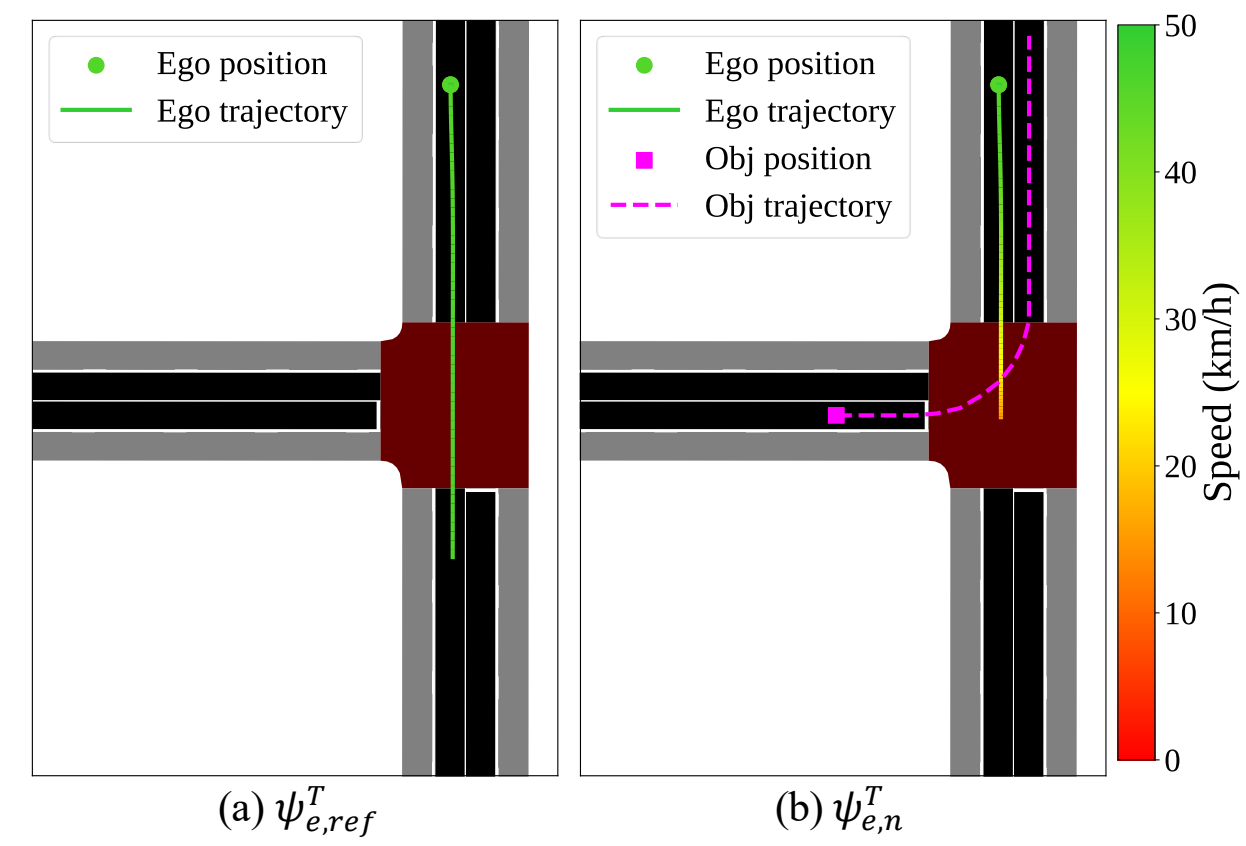

Fig. 3. Counterfactual analysis example. Ego-vehicle with speed $v_e$ = 50 km/h, planning to cross the intersection. Object with ADE$_n$ = 6.1 m.

### A. Relevance Analysis

An illustrative example of the counterfactual analysis performed to determine each object's relevance and encode it within the causal dataset is shown in Fig. 3. In Fig. 3, the Autoware-based ego-vehicle approaches the intersection from the top with speed $v_e$ = 50 km/h, planning to cross it and proceed straight. Note that the color of the ego-vehicle's position and planned trajectory depends on its speed in Fig. 3. Fig. 3(a) depicts the reference trajectory $\psi_{e,ref}^T$ planned by the ego-vehicle in the empty reference scenario with no surrounding objects. Fig. 3(b) depicts the trajectory $\psi_{e,n}^T$ planned by the ego-vehicle in the presence of an object $o_n$ approaching the intersection from the left and planning to turn left, as indicated by its predicted trajectory. The comparison between Fig. 3(a) and Fig. 3(b) reveals that $o_n$ is relevant for the ego-vehicle's driving decisions, as it forces the ego-vehicle to plan a deceleration to avoid a potential collision. In this case, the relevance or measured ADE$_n$ is equal to 6.1 m. In the remainder of this work, we will label as relevant any object with ADE$_n > 0$ m.

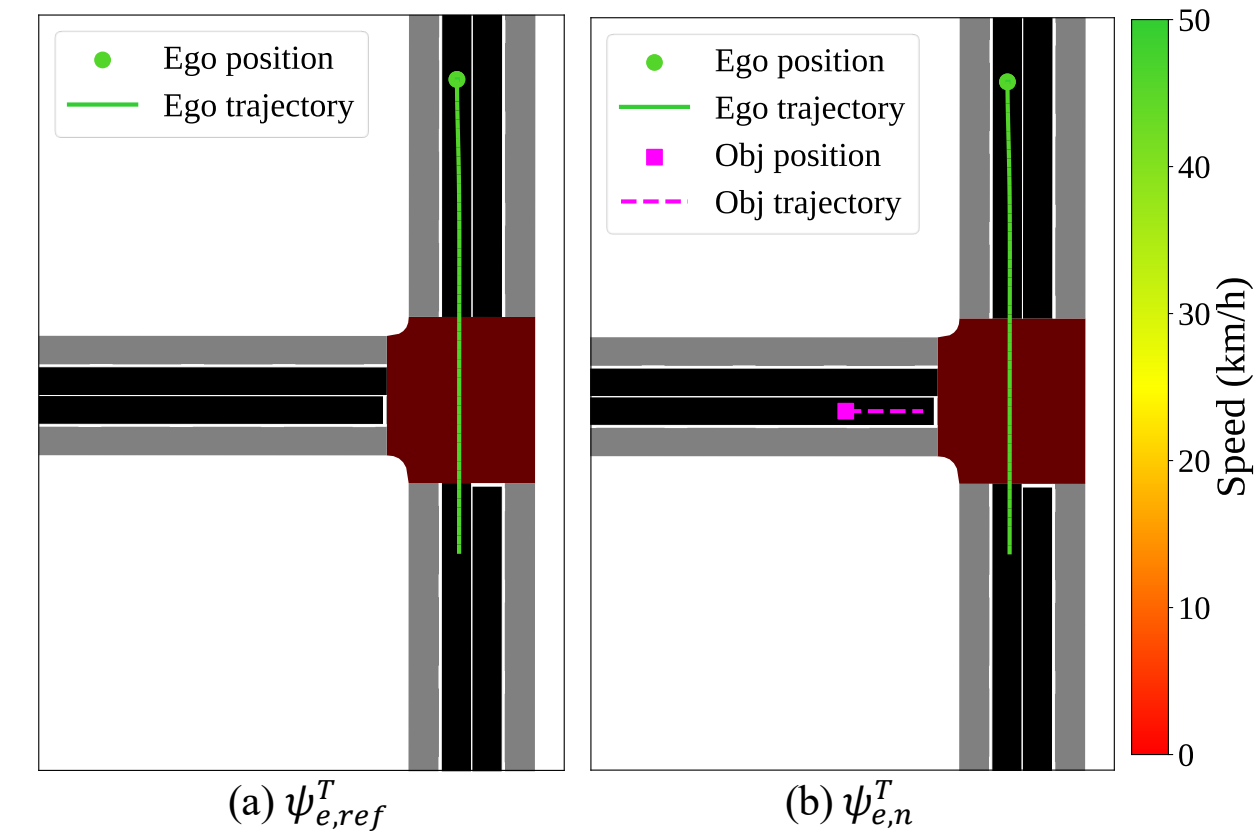

Fig. 4. Counterfactual analysis example. Ego-vehicle with speed $v_e$ = 50 km/h, planning to cross the intersection. Object with ADE$_n$ = 0 m.

The relevance of an object that stops before the intersection and yields to the ego-vehicle is analyzed in Fig. 4. Like in Fig. 3, the ego-vehicle plans to cross the intersection and its speed is $v_e$ = 50 km/h in Fig. 4. The comparison between $\psi^T_{e,ref}$ and $\psi^T_{e,n}$, respectively reported in Fig. 4(a) and Fig. 4(b), reveals that the two planned trajectories now perfectly coincide. The object's $\text{ADE}_n$ is equal to 0 m in this case. This indicates that the object $o_n$ has no impact on the driving decisions of the ego-vehicle and, therefore, it is irrelevant. The comparison between Fig. 3 and Fig. 4 demonstrates that the counterfactual analysis is able to capture the object's relevance and its sensitivity to the object's future trajectory. Note that, under the same ego-vehicle conditions, the object analyzed in Fig. 4 is labeled as relevant in the human-annotated dataset employed in [8]. While a static object stopping before the intersection does not have any impact on the ego-vehicle's driving (and is thus irrelevant), it may still be perceived as a potential danger by a human annotator. By directly measuring an object's impact on the autonomous vehicle's planned trajectory, the proposed approach guarantees a more accurate relevance estimation than existing approaches that rely on human annotations and may suffer from a mismatch between the annotator's common sense and the autonomous vehicle's decision-making.

## B. Relevance Model

The relevance of an object $o_n$ quantifies its impact on the ego-vehicle's driving (i.e., its planned trajectory) and is measured through the $\text{ADE}_n$ (see Sec. II.B). We put forth an AI-based relevance model that aims to learn the causal relationships that influence an object's relevance (e.g., see Figs. 3-4). Its objective is to accurately estimate – or predict – the $\text{ADE}_n$ of any object (i.e., its relevance) based on its mobility (position, heading, speed, predicted trajectory, etc.) and the ego-vehicle's state (position, heading, speed, future intentions). We design the relevance model as a regression task and we train it using the $\text{ADE}_n$ values encoded within the causal dataset as the only supervision signal.

We implement the relevance model on a LightGBM architecture [14]. This design choice aims to achieve high-accuracy predictions while limiting the model size – a key requirement to confine the inference latency and meet real-time relevance estimation requirements. We train the relevance model on a causal dataset that consists of 10000 driving scenarios. Driving scenarios are generated using SUMO and considering vehicular densities ranging from 50 to 400 veh/km$^2$ to represent different traffic conditions. The range of collected $\text{ADE}_n$ values is [0, 10.4] m. We encode the ego-vehicle's position and goal destination and each object's position and predicted trajectory in a polar coordinate system centered on the intersection. The angular component is discretized into three categorical values representing the top, bottom, and left sides of the intersection. Similarly, the ego-vehicle and objects' heading is discretized into four distinct values representing the 45°-135°, 135°-225°, 225°-315°, and 315°-45° ranges. The relevance model was trained using an 80/20 train-test split. Hyperparameters were optimized via Bayesian search, resulting in a LightGBM configuration with 75 leaves, 0.05 learning rate, maximum tree depth equal to 19, and L2 regularization with $\lambda_2$ = 1.7. Training was monitored with Root Mean Square Error (RMSE-)based early stopping over 50 rounds. Model evaluation on the test set yielded excellent results, with mean RMSE = 0.5 m and $R^2$ = 0.79.

The obtained LightGBM-based relevance model achieves an inference latency of 0.9 ms per object on a desktop computer equipped with a single Intel i9-9980XE CPU and 64 GB RAM. A millisecond-level inference latency is key to enable real-time relevance estimation also in high-density scenarios, where the ego-vehicle may detect up to 30 objects with its onboard sensors alone [15]. Note that the number of objects to be analyzed can grow even further if vehicles can exchange their locally detected objects through Vehicle-to-Everything (V2X) communications. Also note that the millisecond-level latency offered by the relevance model is substantially smaller with respect to the latency of existing relevance estimation approaches based on online counterfactual analysis. Online counterfactual analysis estimates in real-time, while the ego-vehicle drives, the objects' relevance through the continuous generation and evaluation of counterfactual scenarios. It requires the ego-vehicle to calculate a new planned trajectory for each analyzed object and, therefore, it is inherently characterized by higher latencies. In Autoware, the calculation of a new trajectory alone requires approximately 160 ms; in [8] and [9], it requires 127 ms and 150 ms, respectively. Unlike the relevance model proposed in this work, which is trained on a causal dataset generated with offline counterfactual analysis, online counterfactual analysis may be unable to satisfy the real-time constraints of autonomous driving, particularly when the number of detected objects is high.

## C. Relevance Heatmaps

An autonomous vehicle can leverage the relevance model derived with the proposed approach to estimate the relevance of detected objects in real time and, for example, filter or de-prioritize those identified as irrelevant to optimize the usage of computational resources. The relevance model can also be employed to build relevance heatmaps. We define relevance heatmaps as 2D grid-based representations of the ego-vehicle's surrounding environment organized in 2 x 2 m$^2$ cells. Each cell reports the maximum estimated relevance that the objects located therein could have for the ego-vehicle. Other cell size and representation criteria are definitely possible. The maximum estimated relevance of a cell is determined by analyzing the relevance of a large set of objects positioned within the cell and with varying speed, heading, and predicted trajectory. Each object's relevance (i.e., $\text{ADE}_n$) is quantified, through the relevance model, considering the ego-vehicle's state when the relevance heatmap is generated.

The relevance heatmap of an autonomous vehicle crossing the intersection from the top with speed $v_e$ = 50 km/h and $v_e$ = 20 km/h is respectively reported in Fig. 5(a) and Fig. 5(b). The comparison between Fig. 5(a) and Fig. 5(b) reveals that the location of the cells containing relevant objects (ADE$_n$ > 0 m) is not affected by the ego-vehicle's speed. Relevant objects are either located ahead of the ego-vehicle, in the same lane, or on other lanes from which they can cross the ego-vehicle's planned trajectory. All objects located behind the ego-vehicle are irrelevant (ADE$_n$ = 0 m). In contrast, the comparison in Fig. 5 shows that the ego-vehicle's speed greatly affects the magnitude of the objects' relevance, as the maximum ADE$_n$ values reported in the cells of Fig. 5(a) are substantially larger compared to their counterparts in Fig. 5(b).

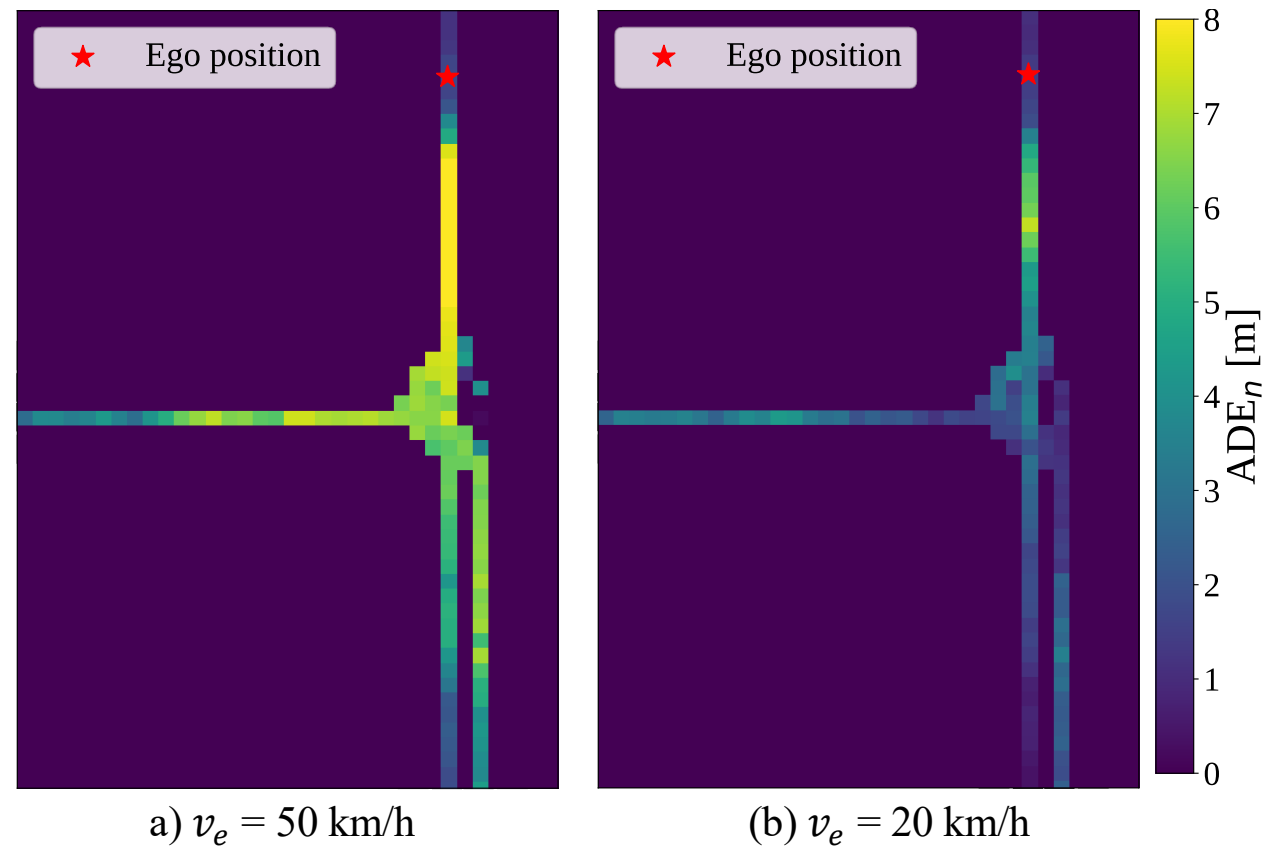

Fig. 5. Relevance heatmap example. Ego-vehicle crossing the intersection from the top at different speeds.

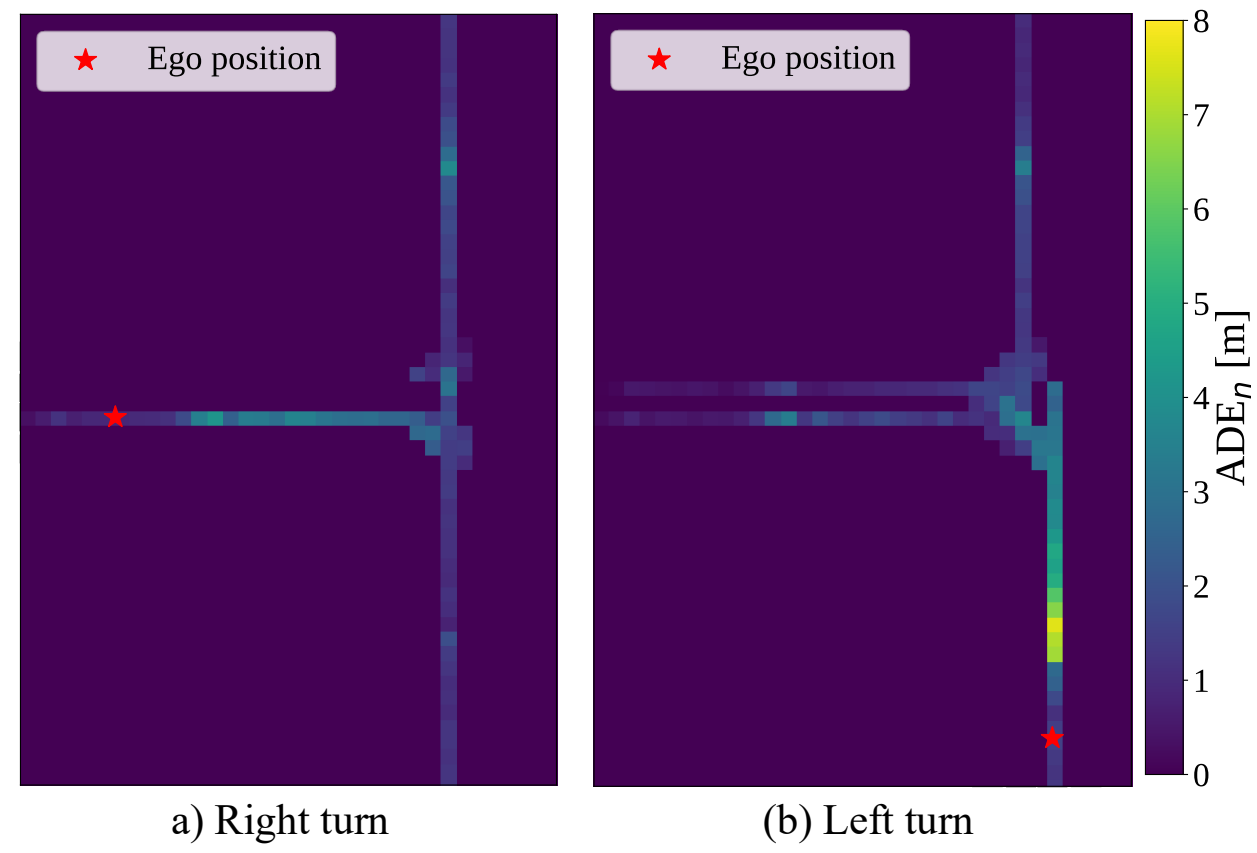

Fig. 6. Relevance heatmap example. Ego-vehicle with speed $v_e$ = 50 km/h and different driving intentions.

Relevance heatmaps obtained considering different ego-vehicle's positions and goal destinations are shown in Fig. 6(a) and Fig. 6(b). In both figures, the ego-vehicle's speed is $v_e$ = 50 km/h. In Fig. 6(a), the ego-vehicle approaches the intersection from the left and plans to turn right. In Fig. 6(b), it approaches the intersection from the bottom and plans to turn left. The relevance heatmap greatly modifies when moving from Fig. 5 to Fig. 6(a)-(b), revealing the sensitivity of the objects' relevance to the ego-vehicle's state – particularly, its driving intentions (turn left/right, go straight). The analysis of Fig. 6(a) and Fig. 6(b) confirms that the cells with relevant objects are located on the lanes where the objects' trajectory can potentially overlap or intersect with the ego-vehicle's planned trajectory. The comparison between Fig. 5(a) and Fig. 6(a)-(b) further highlights the influence of the ego-vehicle's speed on the objects' relevance. In both Fig. 6(a) and Fig. 6(b), the ego-vehicle decelerates and reduces its speed to perform a turn. As a result, the ego-vehicle's relevance heatmap exhibits smaller maximum relevance (ADE$_n$) values compared to Fig. 5(a), where the ego-vehicle does not decelerate.

Relevance heatmaps offer a 2D spatial understanding of the objects' relevance for an autonomous vehicle. In particular, they provide valuable insights into the autonomous vehicle's driving policy by highlighting the location of relevant objects that could affect the ego-vehicle's driving. Relevance heatmaps could be used to proactively and spatially adapt perception and planning efforts. For example, relevance heatmaps could be used to increase the resolution and/or frame rate of sensors monitoring areas of the surrounding environment with high-relevance regions (e.g., ahead of the ego-vehicle) and, conversely, relax the configuration of sensors monitoring areas with no high-relevance regions (e.g., behind the ego-vehicle). With the single-CPU hardware setup employed in this work, an autonomous vehicle could generate accurate relevance heatmaps at 1 Hz (analyzing ~ 1100 objects at 0.9 ms per object) – a sufficiently small value to guide perception and planning efforts in real time.

## IV. Conclusions and Future Work

This paper presented a novel self-supervised approach based on counterfactual analysis to develop an AI-based relevance model capable of estimating the relevance of objects. We shed light on the potential of the proposed approach by: (i) showing that relevance can be more accurately determined compared with human-annotated approaches and that the estimated relevance is not sensitive to confounding effects; and (ii) training a relevance model on a causal dataset generated in a selected urban scenario. Results show that the relevance model can accurately estimate the objects' relevance while outperforming existing relevance estimation approaches in terms of inference latency. On a single CPU, the relevance model is characterized by a millisecond-level inference accuracy that guarantees scalable real-time relevance estimation also in high-density scenarios. We also showed that the relevance model can be leveraged to build relevance heatmaps that provide a spatial understanding of the objects' relevance and can allow autonomous vehicles to spatially adapt perception and planning efforts in real time.

Future work will focus on enhancing the novel approach proposed in this paper in several directions. We will generalize the relevance model to more complex driving scenarios including objects of different types (e.g., vehicles, pedestrians, bikes, etc.). We will explore the integration of the proposed ADE-based relevance definition with safety-related metrics, and we will take into account the objects' mutual influence on each other's relevance. Finally, we will analyse the impact of relevance-aware perception and planning modules on autonomous driving safety and efficiency.